\documentclass[runningheads]{llncs}

\newif\ifarxiv
\arxivtrue

\usepackage[T1]{fontenc}
\usepackage{graphicx}
\usepackage{amsmath}
\usepackage{amssymb}
\usepackage{booktabs}
\usepackage[dvipsnames,table]{xcolor}
\usepackage{tikz}
\usetikzlibrary{arrows.meta,positioning,calc}
\usepackage{multirow}
\usepackage{ifthen}
\usepackage{xspace}
\usepackage[colorlinks=true,allcolors=black]{hyperref}
\usepackage[capitalise]{cleveref} %
\usepackage[acronym]{glossaries}
\glsdisablehyper

\newcommand*{\RL}[2][]{\textcolor{Rhodamine}{[\textbf{\ifthenelse{\equal{#1}{}}{RL}{RL(#1)}}: #2]}}

\newcommand\major[1]{#1} %

\newif\ifblind
\blindfalse    %

\newacronymstyle{long-short-br}
{%
  \GlsUseAcrEntryDispStyle{long-short}%
}%
{%
  \GlsUseAcrStyleDefs{long-short}%
}
\setacronymstyle{long-short-br}

\usepackage{transparent}
\usepackage{tikz}
\ifarxiv
    \newcommand\copyrighttext{%
      \scriptsize Accepted to the BraTS Cluster of Challenges @ MICCAI 2026. The final published version will be available on \textit{Springer}.}
    \newcommand\copyrightnotice{%
    \vspace{-5mm}
    \begin{tikzpicture}[remember picture,overlay]
    \node[anchor=south,yshift=80pt,xshift=0pt] at (current page.south) {\fbox{\transparent{0.85}\parbox{\dimexpr0.625\textwidth-\fboxsep-\fboxrule\relax}{\copyrighttext}}};
    \end{tikzpicture}%
    }
\else
\fi

\begin{document}
\title{Generalizable Brain Tumor Segmentation with Self-Training and Tumor-Aware~Deformations} %
\titlerunning{Self-Training and Tumor-Aware Deformations for BraTS-GoAT}

\ifblind

\author{Anonymous Author(s)}
\authorrunning{Anonymous Author(s)}
\institute{Anonymous Institution\\
\email{anonymous@institution.com}
}
\else

\author{
Henrique Zan Grande\inst{1} \and
Jeovane Honorio Alves\inst{2} \and\\
Rayson Laroca\inst{1}\and
Andre Gustavo Hochuli\inst{1}
}

\authorrunning{H. Z. Grande et al.}

\institute{
Pontifícia Universidade Católica do Paraná (PUCPR), Programa de Pós-Graduação em Informática (PPGIa), Curitiba, Paraná, Brazil\\
\email{\{henrique.zgrande,rayson,aghochuli\}@ppgia.pucpr.br}
\and
University of Luxembourg, SEDAN - SnT, Luxembourg, Luxembourg\\
\email{jeovane.alves@uni.lu}
}
\fi
\maketitle              %

\ifarxiv
    \copyrightnotice
\else
\fi

\newacronym{brats}{BraTS}{Brain Tumor Segmentation}
\newacronym{goat}{BraTS-GoAT}{Generalizability Across Tumors}
\newacronym{hd95}{HD95}{95th percentile Hausdorff Distance}
\newacronym{mri}{MRI}{Magnetic Resonance Imaging}
\newacronym{nsd}{NSD}{Normalized Surface Distance}

\newcommand{\challenge}{BraTS 2026 Challenge\xspace}
\newcommand{\supplementary}{\url{https://github.com/Henrique-zan/brats-goat-2026/}}

\begin{abstract}

This work presents an approach to the \gls*{goat} task of the \challenge, which focuses on robust segmentation of brain tumor sub-regions across a heterogeneous patient population. The proposed method employs the nnU-Net framework with a large residual encoder architecture, integrating a semi-supervised learning technique with pseudo-labels generated from the unlabeled training data and a tumor-aware deformable augmentation that locally deforms the lesion while preserving the surrounding anatomy. We evaluate the individual contributions of each component, as well as their combination, using varying proportions of the most confident pseudo-labeled cases. %
The submitted configuration for the generalization task achieves Dice and NSD scores of 0.881 and 0.473 for Whole Tumor, 0.817 and 0.490 for Tumor Core, and 0.775 and 0.533 for Enhancing Tumor on the \gls*{goat} validation set, improving over the labeled-only baselines across all tumor regions and confirming that self-training and the proposed augmentation are complementary.
Our source code is publicly available at \supplementary.

\end{abstract}
\section{Introduction}
\label{sec:intro}

\setcounter{footnote}{0}
\glsresetall

Brain tumors vary considerably in appearance, size and location, making automatic segmentation from \gls*{mri} images a long-standing challenge.
Accurate delineation of the Whole Tumor, Tumor Core, and Enhancing Tumor is essential for diagnosis, treatment planning, and therapy assessment~\cite{bakas2017advancing,menze2015brats}.
In this vein, the \gls*{brats} challenge has driven advances in this field by providing a large annotated benchmark~\cite{baid2021rsna,bratsgoat2026}.

Most state-of-the-art methods for \gls*{brats} are based on deep encoder-decoder architectures derived from U-Net and its volumetric extension, 3D U-Net~\cite{cicek20163dunet,ronneberger2015unet}.
Building on this architectural family, nnU-Net provides a self-configuring segmentation pipeline that adapts preprocessing, network design, training, and inference to the target dataset~\cite{isensee2021nnunet}.
Its robustness and reproducibility have established it as a reference framework for developing and evaluating medical image segmentation methods \cite{antonelli2022msd,bancerek2025bratsmets}.
Nevertheless, maintaining reliable performance across distinct tumor types, acquisition protocols, imaging sites, and patient populations remains~challenging.

The \gls*{goat} task shifts the emphasis from in-distribution accuracy to generalization across heterogeneous tumor distributions~\cite{bratsgoat2026,karargyris2023medperf}.
Its data include multiple tumor types, acquisition sites, and patient populations, thereby penalizing methods that overfit site-specific image characteristics or a narrow range of tumor morphologies.
This setting motivates the use of unlabeled training data through semi-supervised learning and the introduction of anatomy-aware augmentations that increase morphological diversity without producing unrealistic global deformations~\cite{krinski2023dacov}.
Together, these strategies can broaden the effective training distribution and improve robustness to previously unseen~cases.

To address this generalization challenge, we extend nnU-Net with three complementary components: a region-based teacher built from the Large Residual Encoder preset, a self-training pipeline that uses confident pseudo-labels generated by a modified nnU-Net teacher~\cite{satushe2025ensemble}, and a tumor-aware deformable~augmentation.
The first two components strengthen pseudo-label generation and knowledge transfer, whereas the proposed augmentation constitutes the main methodological contribution by increasing tumor-shape variability while preserving the surrounding~anatomy.

Based on these observations, this work is guided by the following research questions: %

\begin{itemize}
    \item \textbf{RQ1}.  What is the impact of incorporating unlabeled \gls*{goat} data through self-training on segmentation generalization across heterogeneous tumor distributions?
    \item \textbf{RQ2}. How does tumor-aware deformable augmentation complement self-training in improving segmentation robustness and generalization?
\end{itemize}

To answer these questions, we conduct a systematic ablation study in which each component is evaluated independently before being integrated into the complete pipeline. Experiments are conducted using a patient-level holdout to prevent data leakage, and the best-performing arrangement is selected based on validation results from the official \gls*{goat} challenge platform.
The results show that self-training improves overall performance and that the proposed augmentation provides additional gains for Whole Tumor~(WT) and Tumor Core~(TC), while largely preserving the improvements obtained for Enhancing Tumor~(ET).

The remainder of this paper is organized as follows.
\cref{sec:data} describes the challenge dataset and evaluation regions.
\cref{sec:methods} presents the proposed teacher--student framework, tumor-aware augmentation, and training procedure.
\cref{sec:results} reports the quantitative and qualitative results, while \cref{sec:discussion} analyzes the contribution of each component.
Finally, \cref{sec:conclusion} summarizes the main findings and outlines directions for future~work.

\section{Dataset Specification}
\label{sec:data}

We use the official \gls*{goat} dataset from the \challenge~\cite{bratsgoat2026}, a multi-institutional collection designed to evaluate the generalizability of brain tumor segmentation methods across tumor types, acquisition sites, and patient populations. Each case comprises four \gls*{mri} modalities: native T1-weighted~(T1n), contrast-enhanced T1-weighted~(T1c), T2-weighted~(T2w), and T2-FLAIR~(T2f), used as four input channels.
All volumes are skull-stripped, assigned to an anatomical template, and resampled to an isotropic resolution of~$1\mathrm{mm}^3$ ($240\times240\times155$ voxels), following the standard \gls*{brats} preprocessing pipeline~\cite{baid2021rsna,bakas2017advancing}.
Voxel intensities are normalized independently for each modality using nnU-Net default z-score normalization within the non-zero brain~region.

\begin{figure}[!htb]
    \centering
    \includegraphics[width=0.95\linewidth,
    trim={0.25cm 0.25cm 0.25cm 0.05cm}, clip]{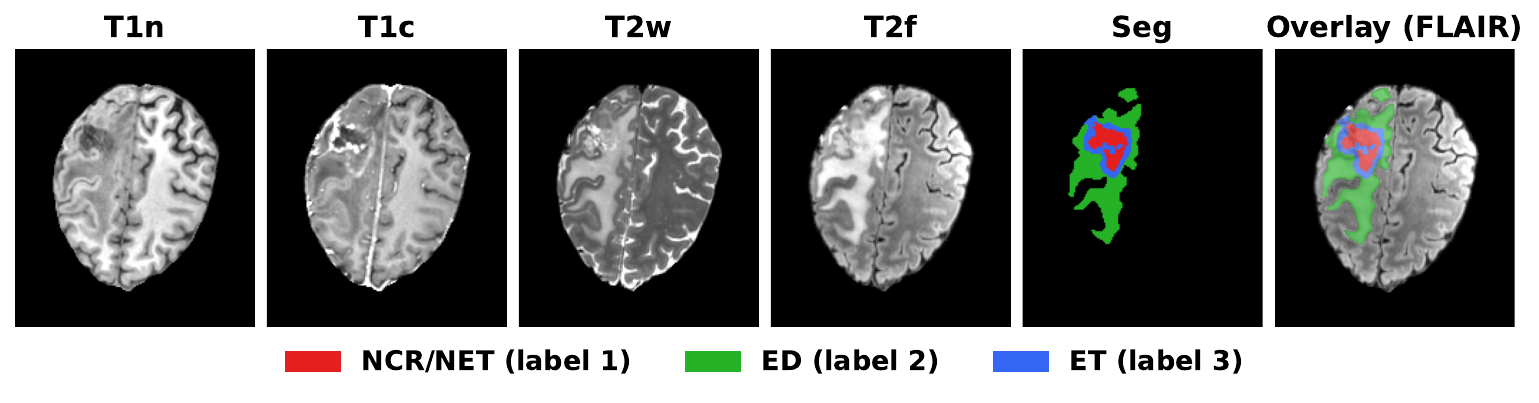}
    
    \vspace{-3mm}
    
    \caption{Example MRI modalities and segmentation masks.
    }
    \label{fig:brats_mrt}
\end{figure}

The dataset comprises a labeled training set, an unlabeled training pool, and an official validation set without public ground truth, evaluated through the challenge platform. The labeled set includes fully annotated adult gliomas~\cite{bakas2017advancing}, partially annotated meningiomas~\cite{labella2023meningioma}, and brain metastases~\cite{moawad2023mets}, while the remaining cases constitute the unlabeled pool. The validation set further includes BraTS-Africa~\cite{adewole2023bratsafrica} and pediatric tumors~\cite{kazerooni2023pediatrics,kazerooni2024pediatrics}, introducing previously unseen tumor types and patient populations. Annotations follow the standard \gls*{brats} protocol: necrotic/non-enhancing tumor core~(NCR/NET, label 1), peritumoral edema~(ED, label 2), and enhancing tumor~(ET, label 3).
Performance is evaluated on the derived regions Whole Tumor (WT $=1\cup2\cup3$), Tumor Core (TC $=1\cup3$), and Enhancing Tumor (ET $=3$). \cref{fig:brats_mrt} shows the four \gls*{mri} modalities and the corresponding segmentation for a representative subject. The dataset contains 1,351 labeled, 1,138 unlabeled, and 451 validation volumes~(patients). %

\section{Method}
\label{sec:methods}

\begin{figure}[!htp]
    \centering
    \includegraphics[width=0.98\linewidth]{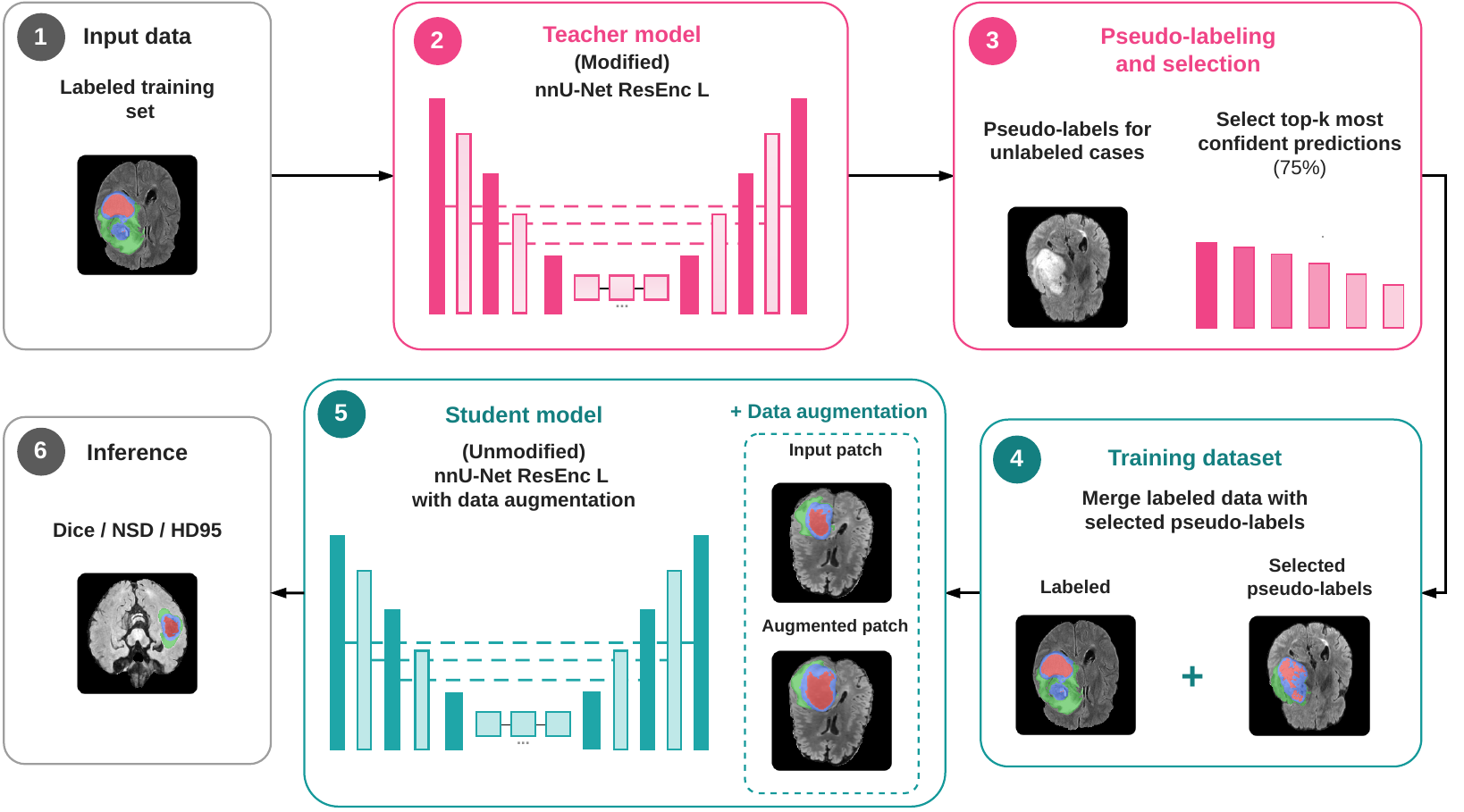}   

    \vspace{-1mm}
    
    \caption{Overview of our nnU-Net-based self-training pipeline for \gls*{goat}.}
    \label{fig:pipeline}
\end{figure}

\cref{fig:pipeline} summarizes the proposed pipeline.
A supervised model~(teacher) is first trained on the labeled data and then used to generate pseudo-labels for the unlabeled cases.
The most confident predictions are merged with the labeled data to improve robustness and generalization.
The second model architecture~(student) %
employs an unmodified large residual encoder nnU-Net~\cite{isensee2024revisited} and is trained with the proposed tumor-aware deformable augmentation to enhance its~representations.

The following sections describe each component in detail.
The complete implementation, including all hyperparameters, source code, and trained model weights, is publicly available for research purposes\footnote{\scriptsize \supplementary}.

\subsection{Supervised Model (Teacher)}
\label{sec:backbone}

To maximize the quality of the pseudo-labels, the teacher model is based on nnU-Net v2~\cite{isensee2021nnunet} using the 3D full-resolution configuration with the Large Residual Encoder preset proposed in~\cite{isensee2024revisited}.
Following Satushe et al.~\cite{satushe2025ensemble}, we further replace the
default instance normalization with group normalization, incorporate axial attention into the decoder, and augment the loss with a Hausdorff-distance term to improve boundary delineation. The model is trained in a region-based manner, predicting the three overlapping tumor regions (WT, TC, and ET) with sigmoid outputs. The teacher is trained with a patient-level five-fold cross-validation, which keeps each patient within a single fold and thus avoids leakage. The final prediction averages the five resulting models to reduce variance and potential~bias.

Although more computationally demanding than the standard nnU-Net, this architecture provides greater capacity to model complex anatomical and morphological variations. Its strong performance in the BraTS 2025 Generalization Challenge~\cite{satushe2025ensemble} further motivated its selection as the teacher model for pseudo-label generation, used in the subsequent training~sequence.

\subsection{Semi-Supervised Model (Student)} \label{sec:ssl}

To leverage the unlabeled training data described in \cref{sec:data}, we adopt a self-training strategy~\cite{lee2013pseudo,xie2020selftraining}, in which a teacher model trained on labeled data generates pseudo-labels that are subsequently incorporated into a new training stage. 

In our approach, the supervised teacher model described in \cref{sec:backbone} predicts the three overlapping tumor regions (WT, TC, and ET). Cases with empty Whole Tumor predictions are discarded, and the remaining pseudo-labeled cases are ranked by confidence, calculated as the mean, across the three regions, of the average foreground probability within each predicted region. The top 75\% most confident cases, corresponding to confidence scores of at least 97.22\%, are retained and merged with the labeled training set. This confidence-based filtering reduces the impact of noisy pseudo-labels while substantially expanding the effective training set.

We selected the retention fraction by comparing 25\%, 50\%, 75\%, and 100\% of
the confidence-ranked cases, each trained identically to the student models
except for an 80\%/20\% split adopted only for this comparison, and evaluated on
the \gls*{goat} validation set. As shown in~\cref{tab:pseudofrac}, the 75\% setting offered the best trade-off between the number of retained cases and label reliability, and was therefore adopted for the final model.

\begin{table}[!htb]
\centering
\setlength{\tabcolsep}{5pt}
\renewcommand{\arraystretch}{1}
\caption{\gls*{goat} validation results for the four pseudo-label retention
fractions, taken after discarding the empty-prediction cases. Dice and \gls*{nsd} are reported as percentages, whereas \gls*{hd95}
is in millimeters. For all metrics, results are presented as mean $\pm$ standard deviation over all validation cases.}
\label{tab:pseudofrac}

\vspace{-1.5mm}

\resizebox{.9\columnwidth}{!}{%
\begin{tabular}{llccc@{\hspace{1.2em}}c}
\toprule
Retention & Metric & WT & TC & ET & Avg\\
\midrule

\multirow{3}{*}{\shortstack[l]{25\%\\268 cases}}
 & Dice ($\uparrow$) & $86.3 \pm 20.8$ & $80.2 \pm 28.1$ & $76.2 \pm \phantom{0}30.9$ & $80.9$\\
 & NSD ($\uparrow$)  & $45.3 \pm 20.4$ & $47.8 \pm 28.3$ & $52.2 \pm \phantom{0}26.6$ & $48.4$\\
 & HD95 ($\downarrow$) & $19.3 \pm 69.5$ & $26.4 \pm 82.2$ & $44.9 \pm 114.6$ & $30.2$\\

\midrule

\multirow{3}{*}{\shortstack[l]{50\%\\536 cases}}
 & Dice ($\uparrow$) & $86.9 \pm 20.0$ & $80.2 \pm 28.0$ & $76.3 \pm \phantom{0}30.7$ & $81.1$\\
 & NSD ($\uparrow$)  & $46.1 \pm 20.3$ & $48.3 \pm 27.9$ & $52.5 \pm \phantom{0}26.4$ & $49.0$\\
 & HD95 ($\downarrow$) & $17.0 \pm 63.4$ & $26.1 \pm 80.7$ & $45.2 \pm 114.2$ & $29.4$\\

\midrule

\multirow{3}{*}{\shortstack[l]{75\%\\803 cases}}
 & Dice ($\uparrow$) & $\mathbf{87.7} \pm 18.5$ & $\mathbf{81.3} \pm 26.5$ & $\mathbf{77.3} \pm \phantom{0}29.8$ & $\mathbf{82.1}$\\
 & NSD ($\uparrow$)  & $\mathbf{46.6} \pm 20.0$ & $\mathbf{49.0} \pm 27.5$ & $\mathbf{53.4} \pm \phantom{0}25.9$ & $\mathbf{49.7}$\\
 & HD95 ($\downarrow$) & $15.3 \pm 58.6$ & $21.8 \pm 71.6$ & $\mathbf{40.9} \pm 108.8$ & $26.0$\\

\midrule

\multirow{3}{*}{\shortstack[l]{100\%\\1{,}071 cases}}
 & Dice ($\uparrow$) & $87.4 \pm 18.6$ & $80.8 \pm 26.7$ & $76.8 \pm \phantom{0}30.0$ & $81.7$\\
 & NSD ($\uparrow$)  & $46.2 \pm 19.9$ & $48.4 \pm 27.4$ & $52.7 \pm \phantom{0}26.0$ & $49.1$\\
 & HD95 ($\downarrow$) & $\mathbf{14.7} \pm 56.3$ & $\mathbf{21.4} \pm 69.9$ & $41.1 \pm 108.9$ & $\mathbf{25.8}$\\
\bottomrule
\end{tabular}%
}
\end{table}

Unlike the teacher model, the student employs the unmodified large residual encoder nnU-Net v2 architecture~\cite{isensee2021nnunet,isensee2024revisited} without the additional architectural modifications introduced for pseudo-label generation. The student is trained on the combined labeled and pseudo-labeled dataset using the proposed tumor-aware deformable augmentation described in \cref{sec:augmentation}. 

Training uses a patient-level 95\%/5\% (training/validation) holdout to prevent data leakage and ensure unbiased model selection. The small validation split maximizes the amount of labeled data available for training, which is particularly beneficial in the semi-supervised setting. Final performance is evaluated independently on the official \gls*{goat} validation set.

This design yields a computationally efficient student while leveraging the knowledge captured by the higher-capacity teacher through self-training. Consequently, the student benefits from reliable pseudo-labels, improving generalization without requiring the same model complexity as the~teacher.

\subsection{Tumor-Aware Deformable Augmentation}
\label{sec:augmentation}
The proposed tumor-aware deformable augmentation increases tumor morphological variability while preserving anatomical characteristics. It combines localized elastic deformations and isotropic scaling to expose the network to a broader range of plausible tumor morphologies, thereby improving~generalization.

The deformation is restricted to the lesion and a narrow surrounding region using a Gaussian-smoothed weighting map derived from the dilated tumor mask, ensuring a smooth transition to the unchanged tissue. For each sampled patch containing a sufficiently large tumor, the transform applies either a smooth elastic deformation or an isotropic scaling around the tumor center of mass, each with equal probability. Smaller lesions are excluded to avoid unrealistic deformations and interpolation artifacts. Image intensities and segmentation labels are warped using the same displacement field, with linear and nearest-neighbor interpolation, respectively, preserving geometric consistency. For computational efficiency, the deformation is computed only within an expanded tumor bounding box, where an edge-fade window enforces zero displacement at the boundaries, preventing boundary artifacts. 

\begin{figure}[!htb]
    \centering
    \includegraphics[width=0.975\linewidth, trim={0.25cm 0.25cm 0.25cm 0.05cm}, clip]{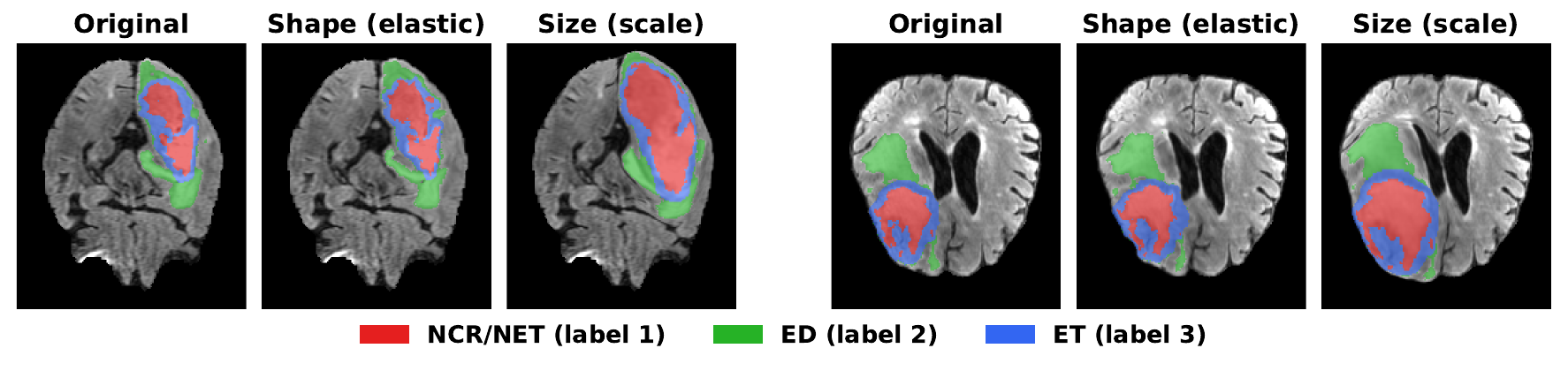}

    \vspace{-3mm}
    
    \caption{Deformable augmentation examples (original, shape, size). %
    }
    \label{fig:augmentation}
\end{figure}

\cref{tab:augparams} summarizes the augmentation parameters, and \cref{fig:augmentation} shows representative examples.

\begin{table}[!htbp]
    \centering
    \caption{Parameters of the tumor-localized deformable augmentation.}
    \label{tab:augparams}

    \vspace{-2mm}

    \resizebox{0.9\linewidth}{!}{    
    \begin{minipage}[t]{0.48\linewidth}
        \centering
        \begin{tabular}{lc}
            \toprule
            \textbf{Parameter} & \textbf{Value} \\
            \midrule
            Application probability
            & $0.5$ \\

            Minimum tumor size (voxels)
            & $50$ \\

            Elastic field smoothness
            & $4.0$ \\

            Maximum elastic displacement
            & $9.0$ \\

            Scale range
            & $0.65$--$1.45$ \\
            \bottomrule
        \end{tabular}
    \end{minipage}
    \hfill
    \begin{minipage}[t]{0.48\linewidth}
        \centering
        \begin{tabular}{lc}
            \toprule
            \textbf{Parameter} & \textbf{Value} \\
            \midrule
            Dilation iterations
            & $8$ \\

            Halo smoothing
            & $5.0$ \\

            ROI margin (voxels)
            & $20$ \\

            Edge-fade width (voxels)
            & $10$ \\

            \phantom{Scale range}
            & \phantom{$0.65$--$1.45$} \\
            \bottomrule
        \end{tabular}
    \end{minipage}
    }
\end{table}

\subsection{Training Procedure}

Both the teacher and the student are trained with stochastic gradient descent,
a polynomial learning-rate schedule, deep supervision, and a combined Dice and
cross-entropy loss. For the students, the patch size~($160\times192\times160$)
and batch size~(3) are automatically determined by the nnU-Net framework using
rule-based heuristics that depend on the dataset and hardware characteristics,
and remain fixed across all student~experiments, ensuring a fair comparison
across all configurations. Each student is also trained for
1000~epochs. The teacher instead uses a fixed patch size~($128\times128\times128$)
and batch size~(5), augments the loss with a Hausdorff-distance term, and is trained
for 250~epochs. For all models, the epoch with the highest
validation Dice score is selected for~inference. All experiments are performed on
an NVIDIA H100 GPU with 80~GB of~memory.

\section{Results}
\label{sec:results}

To fairly evaluate the proposed approach and answer the research questions posed in \cref{sec:intro}, we define an experimental protocol that isolates the contribution of each pipeline component. We first evaluate the supervised teacher model to assess the quality of the generated pseudo-labels. Next, we compare the student trained only on labeled data with the student trained via self-training~(\textbf{RQ1}). We then evaluate the impact of the augmentation alone, applied to the student trained only on labeled data. Finally, we evaluate the complete proposed approach that combines self-training with the tumor-aware deformable augmentation~(\textbf{RQ2}).

As the validation ground truth is not publicly available, predictions are generated locally for the validation cases and submitted to the challenge platform, which returns the official Dice, \gls*{nsd}, and \gls*{hd95} metrics. For each metric, we report the mean and standard deviation across predictions for 451 validation~cases.

\cref{tab:main} summarizes the ablation results.
The first configuration is the supervised teacher evaluated directly on the validation set. The second is the student trained only on labeled data. The third adds confidence-filtered pseudo-labels through self-training. The fourth applies the tumor-aware deformable augmentation to the labeled-only student. The fifth is the submitted model, which combines self-training with the augmentation.

\begin{table}[!htb]
\centering
\setlength{\tabcolsep}{4pt}
\renewcommand{\arraystretch}{1}
\caption{\gls*{goat} validation results for the teacher and the four student
configurations, evaluated through the submission platform. Dice and \gls*{nsd} are
reported as percentages, whereas \gls*{hd95} is in millimeters. For all metrics, results are
presented as mean $\pm$ standard deviation over all validation cases.}
\label{tab:main}

\vspace{-1mm}

\resizebox{.9\columnwidth}{!}{%
\begin{tabular}{llccc@{\hspace{1.2em}}c}
\toprule
Model & Metric & WT & TC & ET & Avg\\
\midrule

\multirow{3}{*}{\shortstack[l]{Teacher\\Labeled Data Only}}
 & Dice ($\uparrow$) & $86.7 \pm 20.3$ & $80.2 \pm 28.4$ & $77.3 \pm \phantom{0}29.8$ & $81.4$\\
 & NSD ($\uparrow$) & $45.5 \pm 20.1$ & $48.3 \pm 27.8$ & $53.0 \pm \phantom{0}25.9$ & $48.9$\\
 & HD95 ($\downarrow$) & $16.9 \pm 63.2$ & $25.0 \pm 79.1$ & $41.0 \pm 109.3$ & $27.6$\\

\midrule

\multirow{3}{*}{\shortstack[l]{Student\\Labeled Data Only}}
 & Dice ($\uparrow$) & $86.6 \pm 20.9$ & $81.1 \pm 27.4$ & $76.0 \pm \phantom{0}31.4$ & $81.2$\\
 & NSD ($\uparrow$) & $46.8 \pm 20.9$ & $48.6 \pm 27.6$ & $52.1 \pm \phantom{0}26.5$ & $49.1$\\
 & HD95 ($\downarrow$) & $19.3 \pm 69.5$ & $25.6 \pm 82.1$ & $46.8 \pm 117.3$ & $30.6$\\

\midrule

\multirow{3}{*}{\shortstack[l]{Student\\w/ Pseudo-Labels}}
 & Dice ($\uparrow$) & $87.8 \pm 18.5$ & $80.3 \pm 27.7$ & $\mathbf{77.5} \pm \phantom{0}29.1$ & $81.9$\\
 & NSD ($\uparrow$) & $46.9 \pm 19.9$ & $48.4 \pm 27.6$ & $53.4 \pm \phantom{0}25.7$ & $49.5$\\
 & HD95 ($\downarrow$) & $15.6 \pm 60.8$ & $23.3 \pm 75.2$ & $\mathbf{39.2} \pm 106.9$ & $26.0$\\

\midrule

\multirow{3}{*}{\shortstack[l]{Student\\w/ Tumor-Aware Aug.}}
 & Dice ($\uparrow$) & $87.3 \pm 19.2$ & $81.5 \pm 26.6$ & $76.9 \pm \phantom{0}30.5$ & $81.9$\\
 & NSD ($\uparrow$) & $\mathbf{47.3} \pm 21.1$ & $\mathbf{49.9} \pm 28.2$ & $\mathbf{53.5} \pm \phantom{0}26.8$ & $\mathbf{50.2}$\\
 & HD95 ($\downarrow$) & $15.2 \pm 58.4$ & $\mathbf{18.7} \pm 65.3$ & $41.9 \pm 111.0$ & $\mathbf{25.3}$\\

\midrule

\multirow{3}{*}{\shortstack[l]{Student (submitted)\\w/ Pseudo-Labels and\phantom{-}\\Tumor-Aware Aug.}}
 & Dice ($\uparrow$) & $\mathbf{88.1} \pm 17.8$ & $\mathbf{81.7} \pm 25.8$ & $\mathbf{77.5} \pm \phantom{0}29.5$ & $\mathbf{82.4}$\\
 & NSD ($\uparrow$) & $\mathbf{47.3} \pm 19.9$ & $49.0 \pm 27.3$ & $53.3 \pm \phantom{0}25.7$ & $49.9$\\
 & HD95 ($\downarrow$) & $\mathbf{14.0} \pm 56.0$ & $21.5 \pm 71.6$ & $40.7 \pm 108.9$ & $25.4$\\
\bottomrule
\end{tabular}%
}
\end{table}

\major{As a complementary analysis, \cref{tab:internalval} reports the mean Dice scores obtained on validation splits derived from the challenge-released training data and used for model selection. The left block corresponds to \cref{tab:main} and includes the five-fold teacher average together with results on the 95\%/5\% student holdout (68 cases), whereas the right block corresponds to \cref{tab:pseudofrac} and reports results using an 80\%/20\% holdout (271 cases). Since the two blocks are based on different validation protocols, their results should be interpreted only through within-block comparisons. Moreover, as these splits are derived from the training data, the reported scores reflect in-distribution performance and should not be interpreted as estimates of generalization to the official challenge validation~set.}

\begin{table}[!htb]
    \centering
    \setlength{\tabcolsep}{5pt}
    \renewcommand{\arraystretch}{1}
    \caption{\major{Internal validation results for the evaluated models and pseudo-label retention fractions, as computed by nnU-Net at the end of training. Dice scores are reported as percentages and averaged across all validation~cases.}}
    \label{tab:internalval}

    \vspace{-4mm}

    \resizebox{0.975\linewidth}{!}{%
        \begin{tabular}[t]{lc}
            \toprule
            Model                       & Avg. Dice ($\uparrow$) \\
            \midrule
            \major{Teacher, Labeled Data Only}  & \major{$89.8$} \\
             \major{Student, Labeled Data Only}  &  \major{$81.3$} \\
             \major{Student w/ Pseudo-Labels}    &  \major{$80.9$} \\
             \major{Student w/ Tumor-Aware Aug.} &  \major{$80.9$} \\
             \major{Student (submitted)}         &  \major{$81.2$} \\
            \bottomrule
        \end{tabular}
        \hspace{1em}
        \begin{tabular}[t]{lc}
            \toprule
            Retention                     & Avg. Dice ($\uparrow$) \\
            \midrule
             \major{25\% pseudo-labels}            &  \major{$82.4$}           \\
             \major{50\% pseudo-labels}            &  \major{$82.0$}           \\
             \major{75\% pseudo-labels}            &  \major{$82.2$}           \\
             \major{100\% pseudo-labels}           &  \major{$82.1$}           \\
            \phantom{100\% pseudo-labels} & \phantom{$82.1$} \\
            \bottomrule
        \end{tabular}%
    }
\end{table}

\section{Discussion}
\label{sec:discussion}

Comparing the supervised teacher with the student trained only on labeled data reveals that the higher-capacity teacher provides only marginal performance differences despite its substantially greater computational complexity. While the teacher achieves slightly better average Dice and \gls*{hd95}, the standard nnU-Net student obtains comparable results, even outperforming the teacher on some individual metrics and tumor regions. This observation suggests that the increased architectural complexity is not justified for direct deployment. Instead, its primary value lies in producing reliable pseudo-labels that transfer knowledge to a simpler and more efficient student through self-training.

Among the student models, incorporating pseudo-labels improves the average Dice score from 81.2\% to 81.9\% while reducing the average HD95 from 30.6 to 26.0~mm, demonstrating that self-training effectively exploits the unlabeled data~(\textbf{RQ1}). The tumor-aware deformable augmentation improves boundary metrics most clearly: applied to the labeled-only student, it achieves the best average NSD~(50.2\%) and HD95~(25.3~mm) among all configurations. Combined with self-training, the submitted model attains the best average Dice~(82.4\%), with average NSD and HD95 of 49.9\% and 25.4~mm, confirming that the augmentation contributes complementary boundary gains~(\textbf{RQ2}).

Region-wise, the largest gains are observed for Whole Tumor, where the complete approach achieves the highest Dice~(88.1\%), NSD~(47.3\%), and the lowest HD95~(14.0~mm). Tumor Core also benefits from the combined strategy, improving all three evaluation metrics over both the baseline and the self-training-only configuration. For Enhancing Tumor, self-training provides the largest improvement, increasing Dice from 76.0\% to 77.5\% and reducing HD95 by more than 7~mm, while the proposed augmentation preserves these gains. Overall, the results indicate that self-training and the proposed tumor-aware augmentation are complementary: pseudo-labels primarily expand the effective training distribution~(\textbf{RQ1}), whereas the augmentation improves robustness to morphological variability, particularly for the larger tumor regions~(\textbf{RQ2}).

\begin{figure}[!ht]
\centering
\includegraphics[width=0.85\linewidth,trim={0.25cm 0.25cm 0.25cm 0.05cm}, clip]{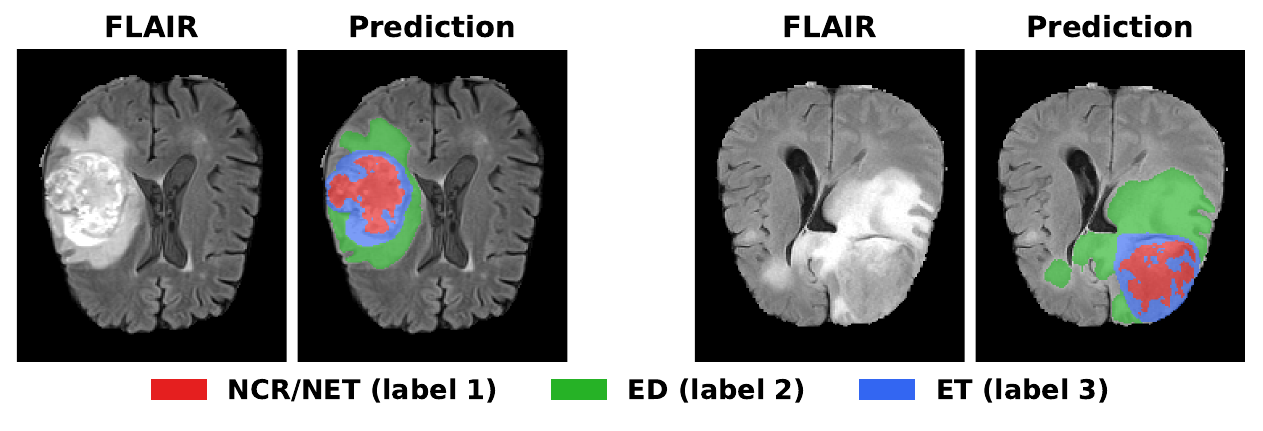}

\vspace{-3mm}

\caption{Qualitative segmentation results of the submitted model on the validation set.
}
\label{fig:qualitative}
\end{figure}

\cref{fig:qualitative} presents predictions from the submitted model for two representative validation cases, overlaid on the FLAIR images.
As the validation annotations are unavailable, these examples support only a qualitative inspection of the predicted NCR/NET, ED, and ET regions, rather than a direct visual comparison with the ground~truth.

\section{Conclusions}
\label{sec:conclusion}

This paper presented an approach to the \gls*{goat} task of the \challenge that combines confidence-filtered self-training with tumor-aware deformable augmentation. Self-training expands the effective training set using high-confidence pseudo-labels, while the proposed augmentation increases the diversity of plausible tumor shapes and sizes without applying global deformations to the surrounding anatomy.

On the official validation set, the combined approach improves both overlap and boundary metrics relative to the labeled-only student, increasing the average Dice score from 81.2\% to 82.4\% and reducing the average \gls*{hd95} from 30.6 to 25.4~mm. The results indicate that the two components are complementary, with the augmentation providing its clearest additional gains for WT and TC. 

\major{Future work will investigate iterative self-training, teacher ensembles, more informative pseudo-label ranking criteria, post-processing strategies, and more detailed analyses across MRI modalities, tumor types, and acquisition domains. We also plan to explore additional methods for further improving segmentation~performance.}

\begin{credits}
\subsubsection{\ackname}

The authors acknowledge the financial support of the \textit{Coordenação de Aperfeiçoamento de Pessoal de Nível Superior}~(CAPES), Finance Code 001; the \textit{Conselho Nacional de Desenvolvimento Científico e Tecnológico}~(CNPq), under Grants No. 302909/2022-2 and 444192/2024-7; the \textit{Financiadora de Estudos e Projetos}~(FINEP), for the High-Performance Computing~(HPC) infrastructure provided by the \textit{Centro Integrado de Soluções em Inteligência Artificial}~(CISIA) at the \textit{Pontifícia Universidade Católica do Paraná}~(PUCPR); and the \textit{Fundação Araucária}, in partnership with the \textit{Secretaria da Ciência, Tecnologia e Ensino Superior do Estado do Paraná}(SETI-PR), under Grant No. 653/2025~(FA/UNIVERSAL).

\subsubsection{\discintname}
The authors have no competing interests to declare that are relevant to the content of this article.
\end{credits}
\bibliographystyle{splncs04}
\bibliography{bibtex}

\end{document}